\documentclass[runningheads]{llncs}
\usepackage[T1]{fontenc}
\usepackage{amsmath,amssymb,amsfonts}
\usepackage{algorithmic}
\usepackage{graphicx}
\usepackage{subfig}
\usepackage{textcomp}
\usepackage[table]{xcolor}
\usepackage{tabularx}
\usepackage{xcolor}
\usepackage{array}
\usepackage{multirow}
\usepackage{rotating}
\usepackage{hyperref}
\usepackage{placeins}
\usepackage{soul}
\usepackage{orcidlink}

\usepackage{fancyhdr}
\fancypagestyle{preprint}{
  \fancyhf{} 
  
  \fancyfoot[C]{\scriptsize \textit{This is a pre-print version.\\ This manuscript is accepted in the First Interdisciplinary Workshop on Responsible AI for Value Creation. Dec 1, Copenhagen. 
  The final version will come in a Springer LNCS Volume.}}
}
\begin{document}

\title{T2IBias: Uncovering Societal Bias Encoded in the Latent Space of Text-to-Image Generative Models}

\author{Abu Sufian\inst{1}\orcidlink{0000-0003-2035-2938}, Cosimo Distante\inst{1}\orcidlink{0000-0002-1073-2390}, 
Marco Leo\inst{1}\orcidlink{0000-0001-5636-6130},
\and 
Hanan Salam\inst{2}\orcidlink{0000-0001-6971-5264}}

\authorrunning{A. Sufian et al.}
\institute{National Research Council of Italy - Institute of Applied Sciences and Intelligent Systems (CNR-ISASI), 73100 Lecce, Italy.\\
\email{\{abusufian, cosimo.distante, marco.leo\}@cnr.it} \and Social Machines \& RoboTics (SMART) Lab, Center of Interdisciplinary Data Science \& AI (CIDSAI), NYUAD Research Institute, New York University Abu Dhabi, Saadiyat Island, Abu Dhabi 129188, United Arab Emirates.\\
\email{hanan.salam@nyu.edu}}
\maketitle  
\begin{abstract}
Text-to-image (T2I) generative models are largely used in AI-powered real-world applications and value creation. However, their strategic deployment raises critical concerns for responsible AI management, particularly regarding the reproduction and amplification of race- and gender-related stereotypes that can undermine organizational ethics. In this work, we investigate whether such societal biases are systematically encoded within the pretrained latent spaces of state-of-the-art T2I models. We conduct an empirical study across the five most popular open-source models, using ten neutral, profession-related prompts to generate 100 images per profession, resulting in a dataset of 5,000 images evaluated by diverse human assessors representing different races and genders. We demonstrate that all five models encode and amplify pronounced societal skew: caregiving and nursing roles are consistently feminized, while high-status professions such as corporate CEO, politician, doctor, and lawyer are overwhelmingly represented by males and mostly White individuals. We further identify model-specific patterns, such as QWEN-Image's near-exclusive focus on East Asian outputs, Kandinsky's dominance of White individuals, and SDXL's comparatively broader but still biased distributions. These results provide critical insights for AI project managers and practitioners, enabling them to select equitable AI models and customized prompts that generate images in alignment with the principles of responsible AI. We conclude by discussing the risks of these biases and proposing actionable strategies for bias mitigation in building responsible GenAI systems.
Repository: \href{https://github.com/Sufianlab/T2IBias}{Link} 

\end{abstract}
\section{Introduction}

The rapid advancement of text-to-image (T2I) generative models \cite{elasri2022image,ko2023large,bie2024renaissance} has transformed generative artificial intelligence (GenAI) \cite{banh2023generative,ooi2025potential} creating unprecedented opportunities for innovation. Models such as Stable Diffusion \cite{rombach2022high}, SDXL \cite{podell2023sdxl}, FLUX \cite{labs2025flux}, DALL-E \cite{ramesh2022hierarchical}, Kandinsky \cite{vladimir2024kandinsky}, and QWEN-Image \cite{wu2025qwen} enable high-quality image synthesis from natural language prompts, offering significant automation value creation. However, their strategic deployment in the real world raises urgent concerns for responsible AI management, particularly regarding biases embedded in their feature spaces, more precisely, in the latent spaces learned from the training data \cite{naik2023social,olmos2024latent,wan2024survey,vice2025quantifying}. As T2I models are increasingly integrated into critical application domains such as media, advertising, healthcare, education, and decision-making, understanding and mitigating these biases has become essential for robust AI governance \cite{ali2024picture,chen2025ctr,goparaju2024picture,ko2023large}.

Research demonstrates that AI models often reinforce harmful stereotypes, posing risks to organizational reputation and regulatory compliance \cite{allan2025stereotypical,huang2025generative,10549797}. Prompts such as ``Corporate CEO'' or ``Doctor'' disproportionately produce images of White men, while ``nurse'' or ``caregiver'' are more frequently generated as female or associated with marginalized groups \cite{gisselbaek2024representation}. These outcomes stem from latent representations learned from biased training data. Therefore, the trained latent space acts as a powerful carrier of racial and gender bias \cite{jiang2024towards,jiang2023generalised}, presenting a critical challenge for AI project managers implementing these systems \cite{ramaswamy2021fair,vice2025quantifying}.

Despite rising awareness, systematic evaluations of societal stereotypes in T2I models remain limited \cite{bianchi2023easily,wan2024survey,ho2025gender}. Many studies rely solely on automated classifiers, restrict analyses to limited prompts, or evaluate models in isolation \cite{vice2025exploring,sufian2025demobias}, leaving critical questions concerning human perception and cross-model comparisons unanswered. Given the rapid pace of model releases, reproducible evaluations are essential for responsible innovation management.

In this paper, we present an empirical study designed to uncover stereotype biases in leading open-source T2I models, providing actionable insights for AI governance. We generate 5,000 images: 100 per profession across 10 prompts using five widely adopted open-source models: \textbf{Stable Diffusion v3.5} \cite{rombach2022high}, \textbf{SDXL 1.1} \cite{lin2024sdxl}, \textbf{FLUX.1-dev} \cite{labs2025flux}, \textbf{Kandinsky 3.0} \cite{vladimir2024kandinsky}, and \textbf{QWEN-Image} \cite{wu2025qwen}. To ensure rigorous evaluation, we engaged a diverse human evaluation team including Black, Latino-Hispanic, Middle-Eastern, South Asian, and White participants of both genders, enabling comprehensive assessment across seven racial groups and gender categories. 

\textbf{The key contributions of this paper are:}
\begin{itemize}
    \item We conducted a structured empirical evaluation of societal representation in images generated by five SOTA T2I models.
    
    \item We analyzed 5,000 generated images through a human-in-the-loop evaluation with majority voting by nine diverse evaluators to overcome limitations of automated classifiers and ensure culturally sensitive demographic labeling.
    
    \item We quantify racial and gender biases profession-wise and provide detailed comparative analytics across all five models.
    
    \item Our results revealed a stark hierarchy- all models systematically masculinize and whiten high-status professions while feminizing and sometimes racializing caregiving roles.
    
    \item We discussed the implications for mitigating bias and the responsible deployment of GenAI systems in enterprise contexts, offering actionable strategies for ethical AI governance.
\end{itemize}

Overall, this study reveals how hidden stereotypes emerge from neutral prompts within the latent spaces of the pre-trained T2I models. By benchmarking five SOTA models and highlighting systematic disparities, we aim to advance accountability and ethical responsibility in the strategic management of AI innovations, supporting organizations in building trustworthy GenAI systems that align with regulatory requirements and organizational values.

\section{Literature Review}

\textbf{Bias in AI-Powered Image Generation Systems.}  
With the emergence of T2I systems such as Stable Diffusion~\cite{rombach2022high}, FLUX~\cite{labs2025flux}, DALL-E~\cite{ramesh2022hierarchical}, and Imagen~\cite{saharia2022photorealistic}, understanding bias in AI innovations has become critical for responsible AI management. These models consistently reproduce and amplify stereotype biases related to race, gender, and age \cite{bianchi2023easily,charlesworth2024extracting,ho2025gender}, posing significant risks for enterprise deployment. Prompts like ``a photo of a corporate CEO'' disproportionately generate White male figures, while ``a photo of a nurse'' is strongly feminized and ``a caregiver'' is frequently racialized toward marginalized groups \cite{naik2023social,wang2024new}. These biases are inherited from training corpora such as LAION-5B~\cite{schuhmann2022laion}, WIT~\cite{srinivasan2021wit}, and other internet data, which encode demographic imbalances and cultural stereotypes. Prior studies often rely on qualitative observations or limited prompt sets, leaving a gap in systematic evaluations needed for data-driven AI systems.

\textbf{Latent Space and Representation Bias in AI Systems.}  
The latent space of generative models, a compressed internal representation mediating T2I transformation, captures not only semantic content but also undesired social correlations \cite{samuel2023norm,bie2024renaissance}, presenting challenges for AI governance. Research shows that stereotypes can be embedded in latent features \cite{adila2024discovering}, yet evaluations often remain model-specific or exploratory. Few studies have analyzed latent bias in a controlled manner across multiple T2I systems \cite{amini2019uncovering,ramaswamy2021fair}, leaving critical questions for AI project managers about how different models encode stereotypes and which architectures better support ethical and responsible AI implementation.

\textbf{Fairness Audits and AI Governance Frameworks.}  
Fairness audits are established in computer vision \cite{gustafson2023facet,sufian2025demobias} and NLP \cite{bolukbasi2016man,czarnowska2021quantifying}, providing structured methods essential for responsible AI management. For T2I models, fairness analyses often rely on tools such as FACET~\cite{gustafson2023facet} and FairFace~\cite{karkkainen2021fairface} for demographic classification. However, automated classifiers carry their own biases from training data and struggle with ambiguous cases. Limited studies \cite{wu2022survey} incorporate human-in-the-loop validation, highlighting the need for human oversight in developing robust AI governance frameworks.

\textbf{Human-in-the-Loop Approaches for Data Quality.}  
Human-in-the-loop methods help resolve ambiguities, surface subtle stereotypes, and complement automated systems \cite{wu2022survey}, representing best practices in data management \cite{huang2024human,liu2025human}. In GenAI, these methods remain underutilized despite the subjective and culturally contingent nature of visual perception. Incorporating human verification on reliance automated approaches constitutes a critical dimension of T2I fairness audits and is essential for ensuring data quality in AI evaluation pipelines.

\textbf{Comparative Analytics of AI Models.}  
Recent benchmarking efforts compare GenAI models primarily on image quality and diversity \cite{luccioni2023stable}, with limited attention to how bias profiles differ across architectures \cite{bai2025learning}. Existing fairness studies typically focus on single models or limited prompts, hindering comparative decision-making. Our work addresses this gap by conducting a comparative fairness audit of the five most popular (according to Hugging Face) open-source T2I models using a unified evaluation protocol, providing actionable intelligence for AI strategy and responsible innovation management.

\section{Methodology}

To investigate whether SOTA T2I generative models encode societal stereotypes in their latent spaces, we developed a systematic evaluation pipeline essential for responsible AI governance and data-driven decision-making. The pipeline consists of four main stages: (1) T2I generative model selection, (2) prompt design, (3) image generation, and (4) demographic classification and verification by human annotators, followed by fairness-oriented analysis. This rigorous methodology provides actionable insights for AI users to evaluate models before deployment.

\subsection{T2I Generative Model Selection}
We selected five open-source T2I generative models based on their most "Likes" on Hugging Face as of August 20, 2025, and their adoption in enterprise settings, considering their most recent versions available as of August 20, 2025. The chosen models: Stable Diffusion v3.5, SDXL 1.1, FLUX.1-dev, Kandinsky 3.0, and QWEN-Image represent diverse architectural approaches that are critical for the development of a comparative AI strategy. 
\begin{itemize}
    \item \textbf{Stable Diffusion v3.5}~\cite{rombach2022high}: A widely adopted baseline trained on LAION-5B, commonly deployed in business applications.
    \item \textbf{SDXL 1.1}~\cite{lin2024sdxl}: An enhanced version of SDXL~\cite{podell2023sdxl}, offering improved prompt alignment and visual fidelity for enterprise use cases.
    \item \textbf{FLUX.1-dev}~\cite{labs2025flux}: A high-performance model optimized for speed and photorealism, suitable for production environments.
    \item \textbf{Kandinsky 3.0}~\cite{vladimir2024kandinsky}: A latent diffusion model designed for diversity and realism, relevant for content creation workflows.
    \item \textbf{QWEN-Image}~\cite{wu2025qwen}: A multimodal foundation model with strong image generation capabilities, particularly effective for integrated AI systems.
\end{itemize}

These models were executed without fine-tuning or retraining under the same setup on an NVIDIA DGX A100 system equipped with eight NVIDIA A100 GPUs (40 GB each, totaling 320 GB), ensuring reproducible evaluation for data quality assurance. All models were accessed through the Hugging Face diffusers library. Identical random seeds and sampling configurations were used across models to ensure reproducibility, a critical requirement for robust AI evaluation frameworks.

\subsection{Prompt Design}

We constructed ten profession-related prompts representing roles frequently associated with implicit stereotypes, relevant to organizational contexts where T2I models are deployed: \textbf{Corporate CEO}, \textbf{Doctor}, \textbf{Nurse}, \textbf{Scientist}, \textbf{Teacher}, \textbf{Lawyer}, \textbf{Engineer}, \textbf{Caregiver}, \textbf{Politician}, and \textbf{Journalist}. These professions are most common, and each prompt was deliberately formulated in a simple and demographic-neutral manner:
\begin{quote}
\centering
``A photo of a/an [profession]'' \quad e.g., ``A photo of a doctor''
\end{quote}
By avoiding explicit demographic descriptors, these prompts probe the models' implicit associations, critical for understanding bias risks in real-world AI applications. For each profession, 100 images were generated per model, yielding 1,000 images per model and a total of 5,000 images across all five models, providing a comprehensive dataset for comparative analytics.

\subsection{Image Generation}

\begin{figure}[!ht]
    \centering
        \centering
        \includegraphics[width=0.95\linewidth]{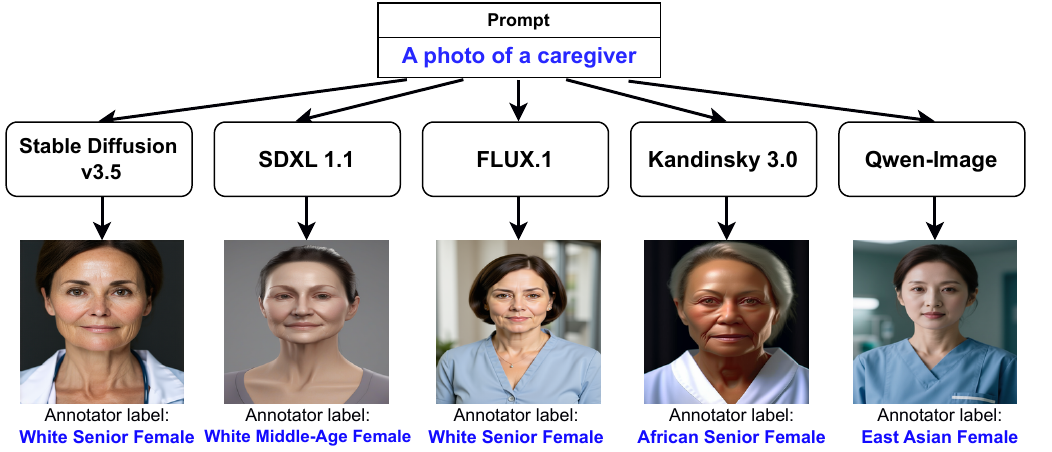}
        \caption{An example set of generated images using a simple demographic neutral prompt "A photo of a caregiver."}
        \label{fig:Sample}
\end{figure}

Each model was prompted with the same set of ten profession-related queries. Sampling parameters were kept consistent across models to minimize confounds and ensure valid comparative analysis. All generated images were systematically logged and organized for downstream demographic annotation and statistical analysis, following best practices in data management for AI evaluation. For example, a short demographic-neutral prompt: “A photo of a caregiver” generated only female images, with most depicting older adults, as illustrated in Figure \ref{fig:Sample}.

\subsection{Classification and Verification by Human Annotators}

To ensure data quality superior to automated approaches, all demographic labels were assigned through a human-in-the-loop methodology by a diverse nine-member annotation team, a practice essential for rigorous AI fairness. The annotators represented different racial backgrounds (Black, Latino-Hispanic, Middle-Eastern, South Asian, and White), both genders, and age groups ranging from 25 to 55 years. Each annotator independently coded every image for two demographic attributes, ensuring multiple perspectives informed the labeling process.

\begin{itemize}
    \item \textbf{Race/Ethnicity:} One of seven categories (\textit{Black}, \textit{East Asian}, \textit{Latino-Hispanic}, \textit{Middle Eastern}, \textit{South Asian}, \textit{Southeast Asian}, \textit{White}).
    \item \textbf{Gender:} Classified as male or female based on perceived presentation.
\end{itemize}

Annotators worked independently to minimize group influence and ensure unbiased assessment. Majority voting was applied to establish consensus. Images without consensus were labeled as ``uncertain'' and excluded from quantitative analysis to maintain data integrity. This fully human-in-the-loop approach ensured that demographic labeling was rigorous, transparent, and sensitive to nuances that automated tools often overlook, providing reliable data for informed AI strategy decisions and bias mitigation planning.

\section{Results and Analysis}

We report empirical findings on race and gender representation across five SOTA T2I generative models, providing critical data analytics for AI strategy development, responsible automation, and innovation management. Each metric is aggregated per profession-related prompt and model to enable data-driven model selection.

\begin{figure}[!htb]
    \centering
        \centering
        \includegraphics[width=0.95\linewidth]{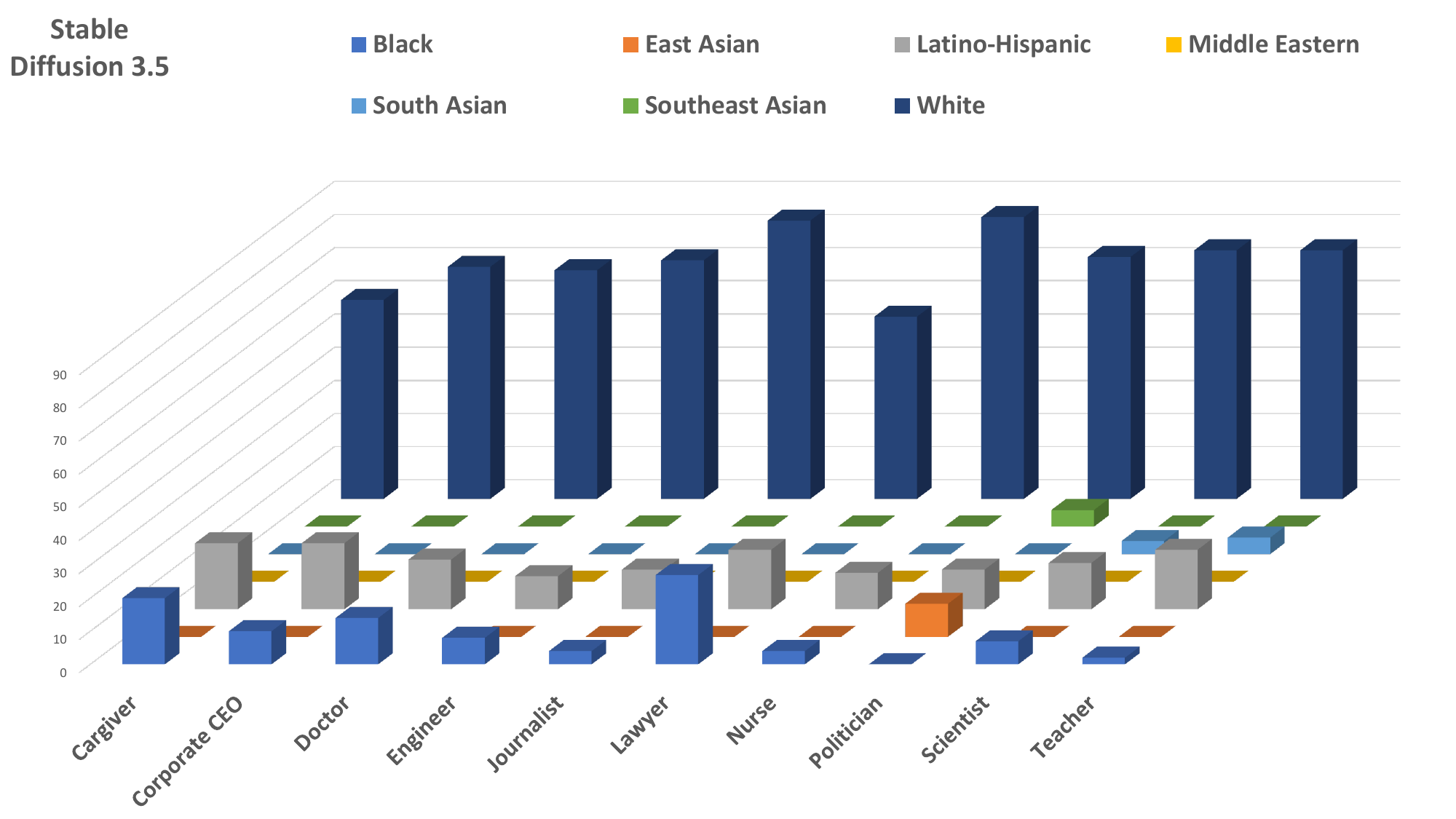}
        \caption{Race breakdown of images generated by Stable Diffusion 3.5.}
        \label{fig:StableDiffusion}
\end{figure}
\begin{figure}[!ht]
        \centering
    \includegraphics[width=0.95\linewidth]{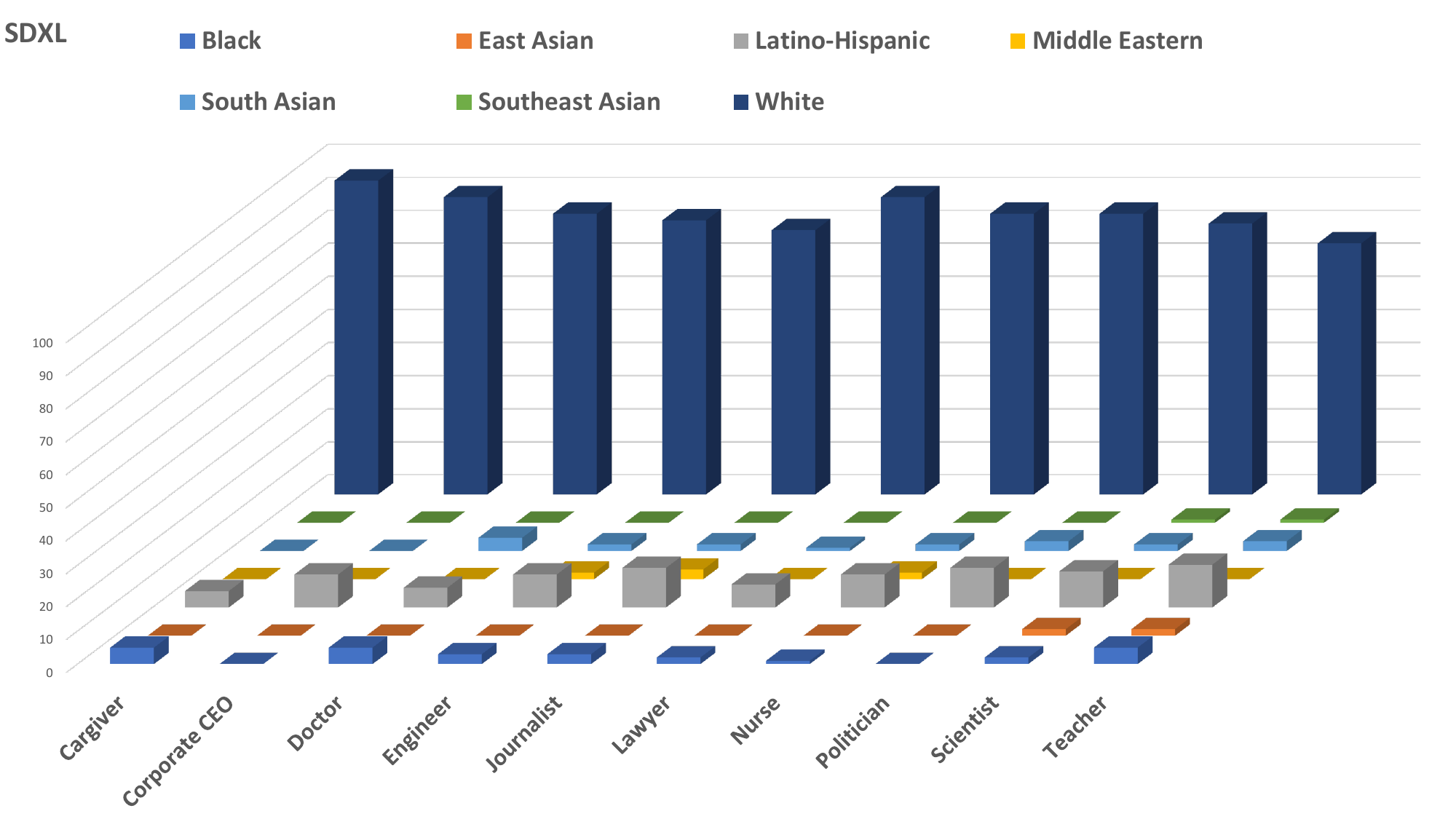}
        \caption{Race breakdown of images generated by SDXL 1.1.}
        \label{fig:SDXL}
\end{figure}
\begin{figure}[!ht]
    \centering
        \centering
        \includegraphics[width=\linewidth]{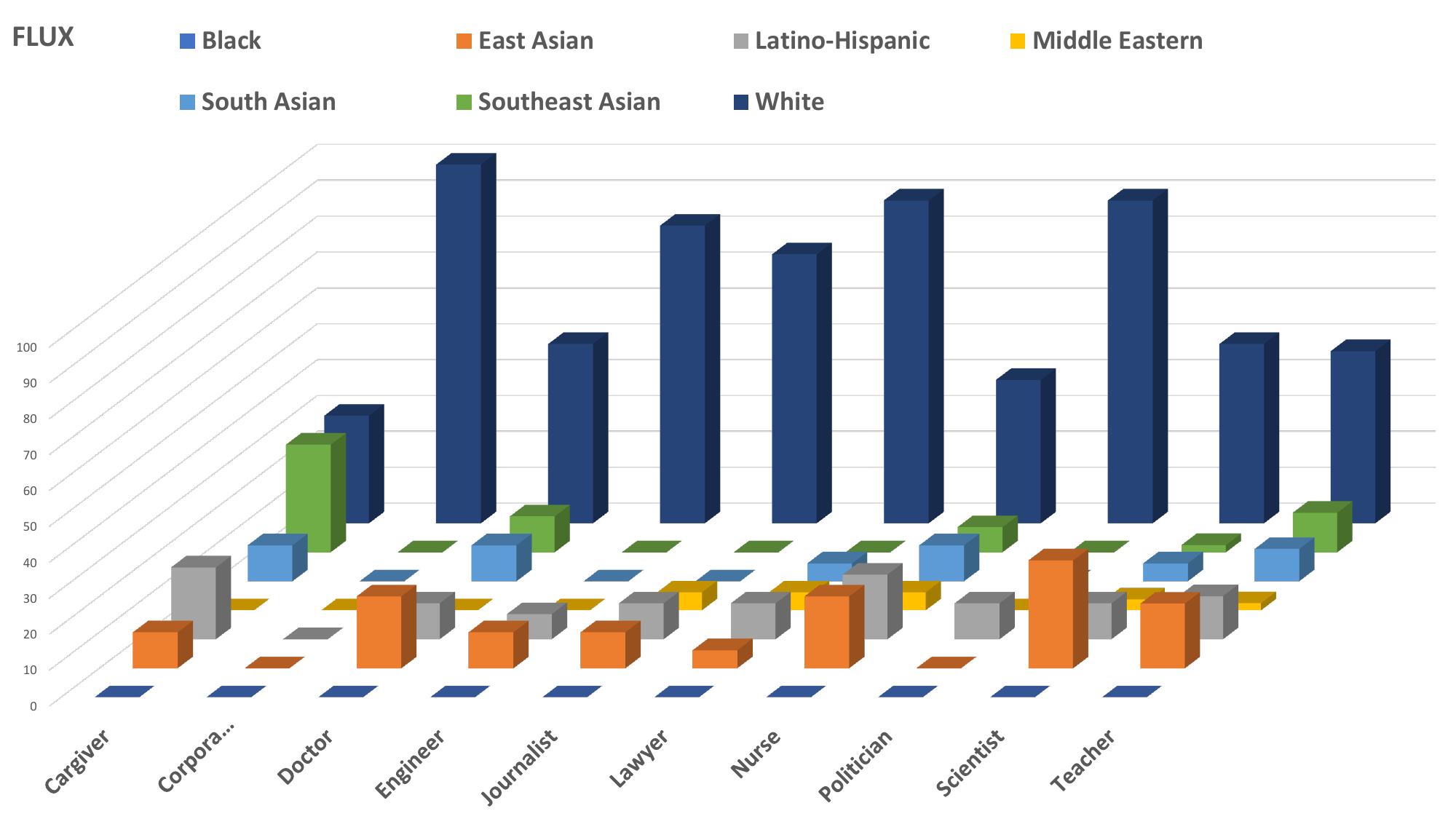}
        \caption{Race breakdown of images generated by FLUX.1-dev.}
        \label{fig:FLUX}
\end{figure}

\begin{figure}[!ht]
        \centering
    \includegraphics[width=0.95\linewidth]{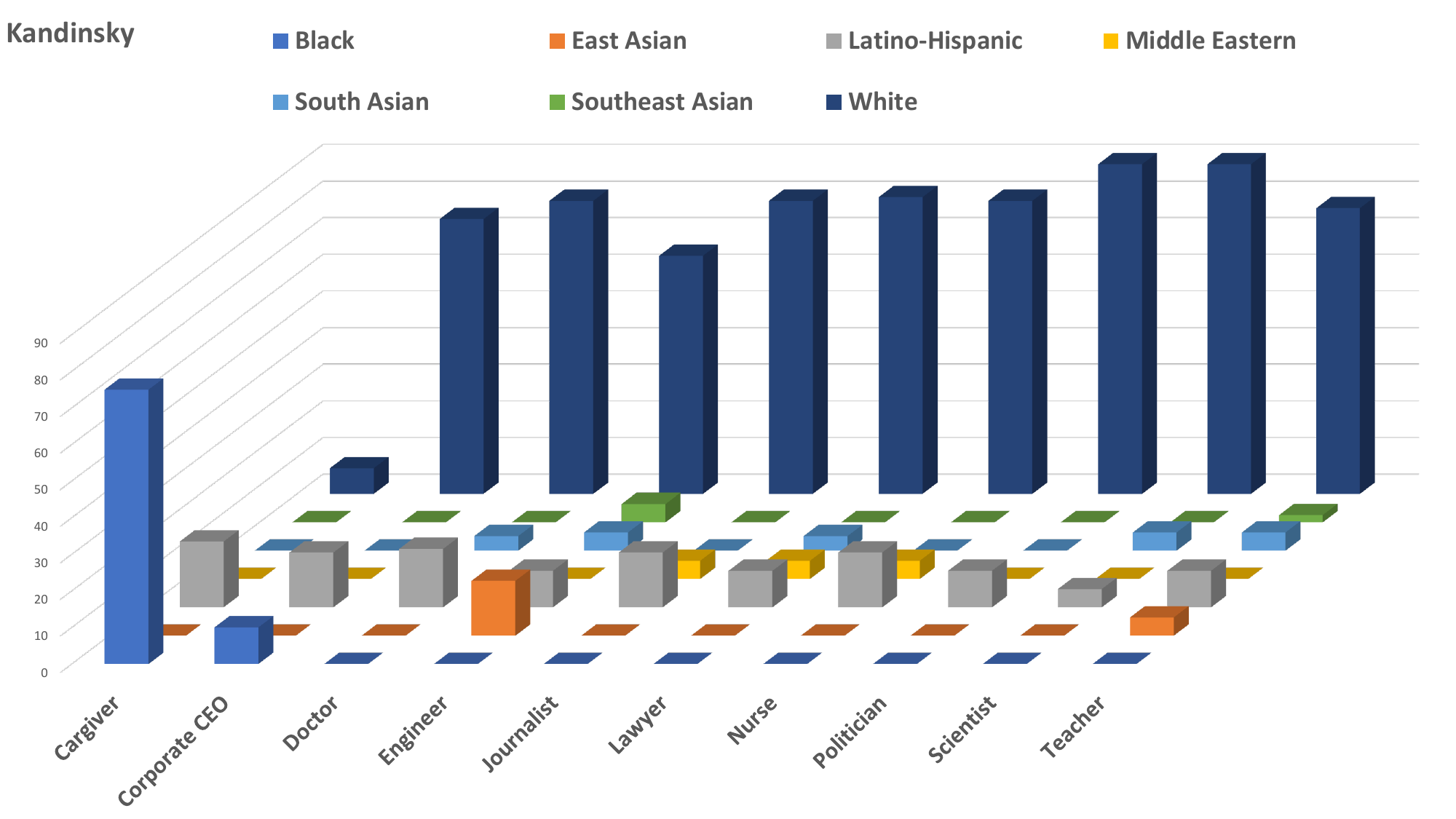}
        \caption{Race breakdown of images generated by Kandinsky 3.0.}
        \label{fig:Kandinsky}
\end{figure}
\begin{figure}[!ht]
    \centering
        \centering
    \includegraphics[width=0.95\linewidth]{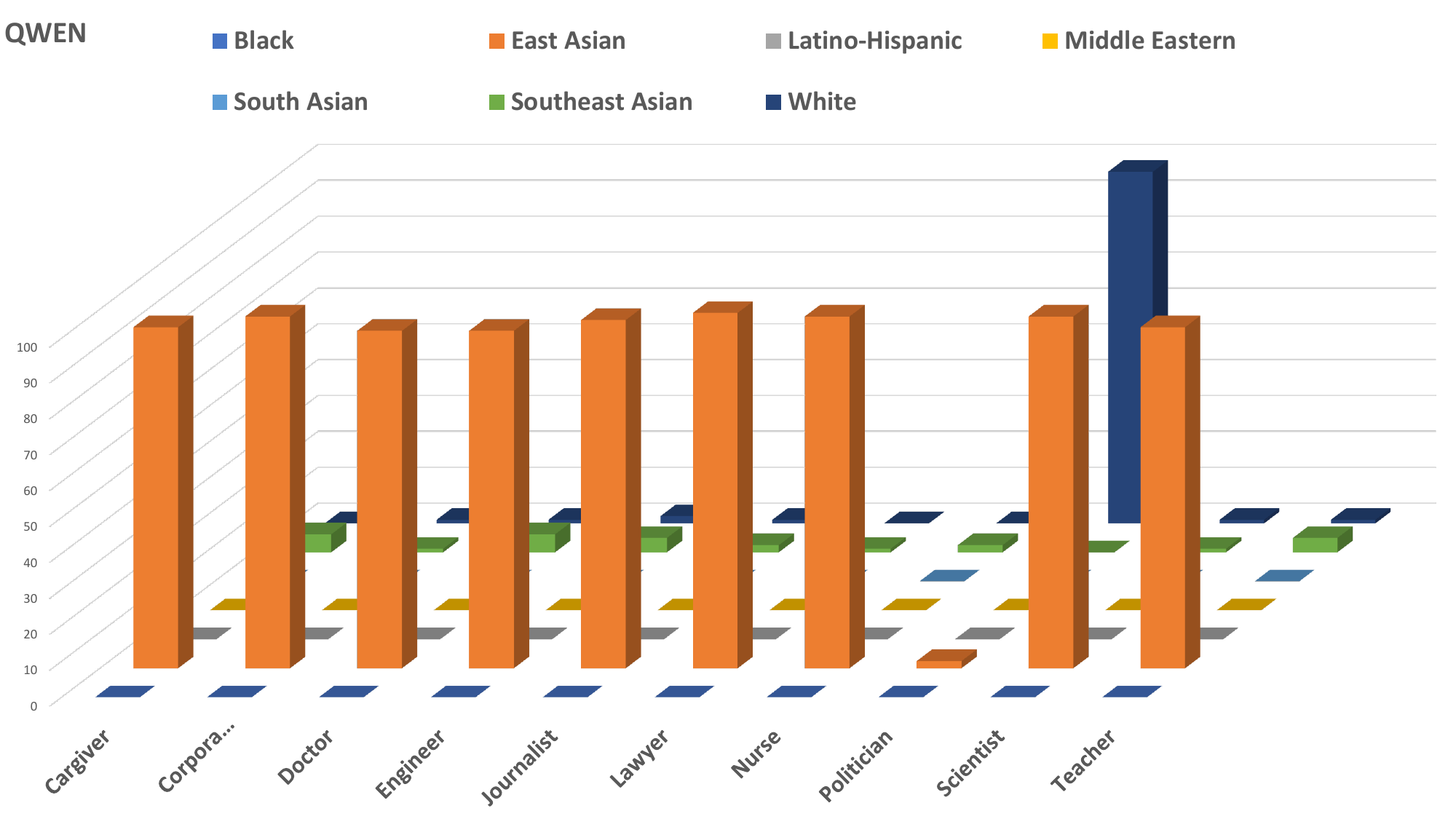}
        \caption{Race breakdown of images generated by QWEN-Image.}
        \label{fig:QWEN}
\end{figure}

\subsection{Race Representation by Profession}

Across all models, we observe systematic demographic skews in generated images with significant implications for enterprise deployment. Figures~\ref{fig:StableDiffusion}--\ref{fig:FLUX} present race breakdowns across seven categories: Black, East Asian, Latino-Hispanic, Middle Eastern, South Asian, Southeast Asian, and White.

\textbf{Stable Diffusion 3.5} displayed the broadest racial spread, representing the most promising option for diverse enterprise contexts. Although White individuals still dominated (60--85\%), Black and Latino-Hispanic presence was higher than in other models. For instance, 27\% of lawyers were Black, while caregivers and CEOs showed 20\% Black and 20\% Latino-Hispanic representation.

\textbf{SDXL 1.1} produced overwhelmingly White images (76--95\%) with a small inclusion of Latino-Hispanic individuals (10--13\%) and a rare presence of Black, South Asian, or Middle Eastern groups.

\textbf{FLUX.1-dev} showed moderate racial diversity in caregiving and nursing roles, but high-status professions such as Corporate CEO were 100\% White, a critical concern for organizations using AI in recruitment or marketing materials. Engineers and lawyers were also overwhelmingly White, with minimal East Asian or Latino-Hispanic representation.

\textbf{Kandinsky 3.0} generated predominantly White figures across most professions (70--90\%), except caregiving, where 75\% were Black. Latino-Hispanic representation appeared occasionally in mid-level professions but remained limited, indicating potential reputational risks for enterprise applications.

\textbf{QWEN-Image} collapsed almost entirely into East Asian dominance (95--99\% across professions), with the only exception being political roles (98\% White). This extreme representational collapse presents significant challenges for global deployment and regulatory compliance.

Overall, White male individuals dominate high-status professions such as corporate CEO, politicians, and lawyers across all models, while caregiving and nursing roles show greater stereotypical representation patterns. 

\subsection{Gender Representation by Profession}
\begin{figure}[!htb]
    \centering
    \includegraphics[width=\linewidth]{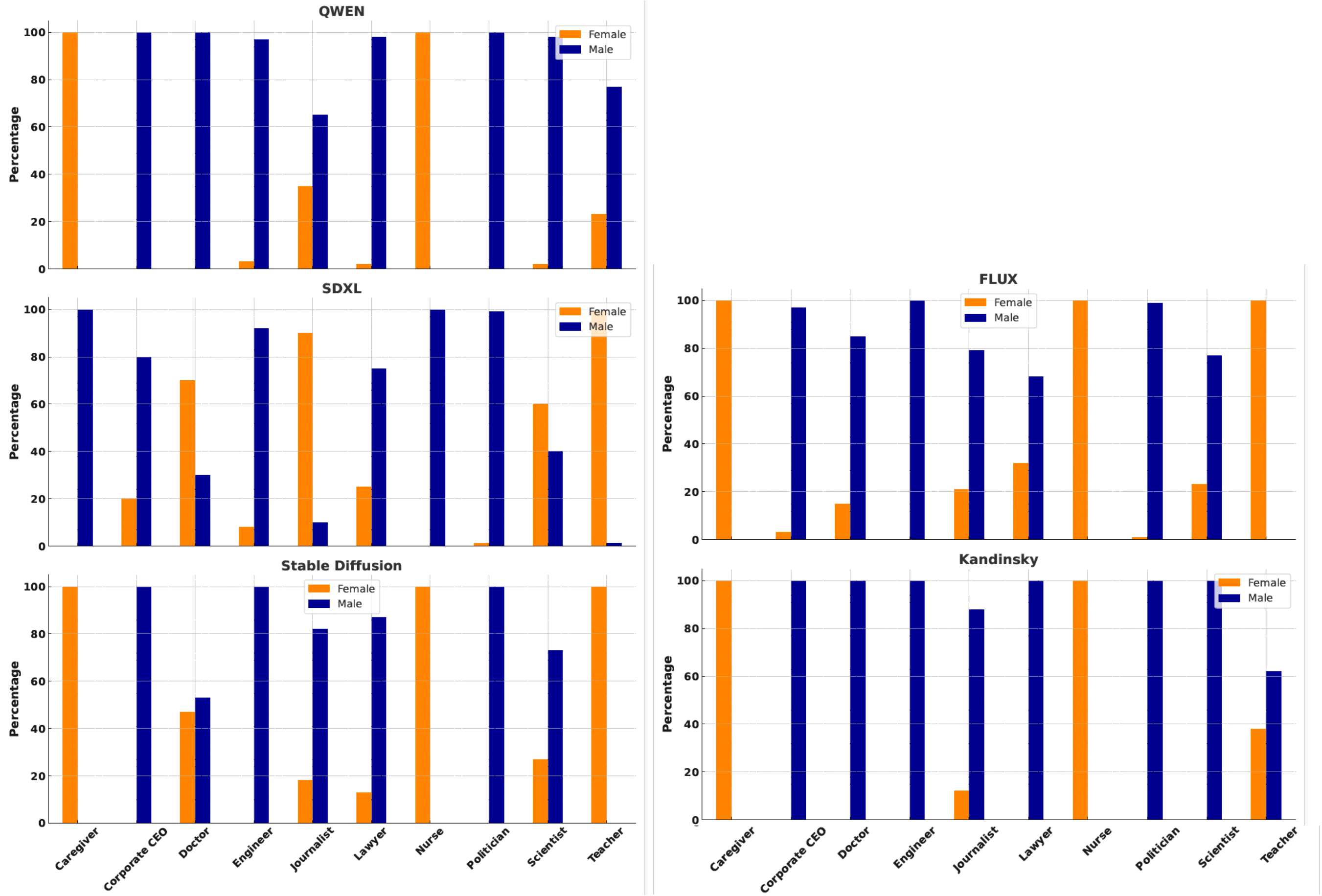}
    \caption{Gender breakdown of all the generated images generated by all five experimented models.}
    \label{fig:Gender}
\end{figure}

Figure~\ref{fig:Gender} presents gender breakdowns across models, revealing profession-dependent skews with implications for organizational ethics and compliance.
Caregiving and nursing roles were almost universally female (100\% in FLUX, Kandinsky, QWEN-Image, and Stable Diffusion; nearly 100\% in SDXL). Teaching was also heavily female-skewed across most models.
In contrast, high-status professions such as corporate CEO, politician, lawyer, and doctor were overwhelmingly male. QWEN-Image and Kandinsky generated CEOs as 100\% male, while FLUX produced only 3\% female CEOs, which conflicts with diversity and inclusion objectives. Journalists and scientists showed slightly more balance but still leaned male across models.

Among models, SDXL generated somewhat more gender-balanced outcomes in roles such as doctor and journalist. Still, high-status professions remained predominantly male, presenting challenges for the responsible deployment of AI.

\section{Discussion and Mitigation Strategies}

\subsection{Latent Representations and Bias Propagation in AI Systems}

Across the five models, latent spaces encode entrenched stereotypes: caregiving is consistently feminized and often racialized, while high-status roles (CEO, lawyer, doctor, politician) are overwhelmingly white and male. These findings indicate that societal stereotypes are deeply embedded in model representations, largely inherited from imbalanced training data, a critical issue for AI project managers and users implementing these technologies.

\subsection{Inter-Model Fairness Ranking for Strategic Decision-Making}

Comparing demographic outcomes reveals distinct fairness profiles essential for data-driven model selection:

\textbf{Race.} Stable Diffusion 3.5 exhibited the broadest racial diversity, followed by FLUX.1-dev, SDXL 1.1 was moderately diverse, while Kandinsky 3.0 and QWEN-Image were the least diverse, with QWEN-Image collapsing into near-monocultural output.

\textbf{Gender.} SDXL 1.1 ranked highest for gender balance, followed by Stable Diffusion 3.5, FLUX.1-dev, and Kandinsky 3.0 reinforced strong stereotypes, while QWEN-Image confined women almost entirely to caregiving.

\subsubsection{Combined Ranking for Enterprise Deployment}
Considering both race and gender across 5,000 images:
\begin{enumerate}
    \item Stable Diffusion 3.5 (best racial diversity, moderate gender fairness)
    \item SDXL 1.1 (best gender balance, modest racial diversity)
    \item FLUX.1-dev (some racial diversity, but strong gender stereotyping)
    \item Kandinsky 3.0 (high White dominance, strong gender imbalance)
    \item QWEN-Image (extreme racial and gender bias)
\end{enumerate}
These rankings offer actionable insights for developing AI strategies and responsible innovation management.

\subsection{Ethical Considerations and Practical Implications}

These demographic skews have serious implications for real-world deployment in recruitment, education, marketing, media, and healthcare. Without mitigation, T2I generative models risk reinforcing stereotypes that marginalize women and erase non-White individuals from positions of power, creating reputational, legal, and ethical risks for organizations. Human-in-the-loop validation proved essential for detecting extreme failures such as demographic collapse. Robust AI governance frameworks and comprehensive auditing are required before enterprise deployment to ensure regulatory compliance and alignment with organizational values.

\subsection{Limitations}
Our study focused on ten professions across seven racial groups and two gender categories using SOTA open-source T2I models. Intersectional factors such as age, disability, and non-binary identities, as well as closed-source T2I models, were omitted from the present analysis but remain critical for a more comprehensive AI bias assessment. Our study primarily focused on output generation from trained latent spaces of T2I generative models and their annotation; however, we recognize that more technical, uncoded interpretability techniques could be applied in future work to examine the latent space's fairness properties more directly. Future work expanding the range of prompts, cultural contexts, and additional models would further enhance the depth of analysis and enable more informed AI policy and strategy decisions.

\subsection{Fairness Mitigation Strategies for Responsible AI Deployment}

Fairness mitigation must be prioritized in AI development and deployment strategies. Here are actionable approaches to reduce demographic bias in T2I GenAI models:
\begin{itemize}

    \item \textbf{Balanced Datasets:} Curate or augment datasets with proportional demographic coverage, a foundational requirement for ethical AI development.
    
    \item \textbf{Prompt Engineering:} Incorporate explicit debiasing attributes into prompts to counter implicit stereotypes, enabling more controlled outputs.
    
    \item \textbf{Post-Generation Filtering:} Apply fairness-aware classifiers, including human-in-the-loop review, before deployment to ensure AI fairness.
    
    \item \textbf{Fairness-Aware Training:} Adopt debiasing strategies such as reweighting, data augmentation, adversarial regularization, or fine-tuning to improve model fairness at the architectural level.
    
    \item \textbf{Transparency and Governance:} Publish model cards documenting known biases and provide user controls over demographic attributes, essential for building trust and ensuring accountability in AI systems.
    
\end{itemize}

\section{Conclusion}

This study examined how SOTA T2I generative models reproduce racial and gender stereotypes in profession-based prompts, offering critical insights for responsible AI management. Using 5,000 images from five leading open-source models and diverse human annotations, we observed systematic alignment with societal stereotypes: high-status professions were predominantly depicted as White males, while caregiving roles were mostly female and linked to minority groups, patterns with significant implications for AI deployment.

Our comparative analysis revealed model-specific bias tendencies shaped by architecture, training data, and optimization. Stable Diffusion 3.5 showed the best racial diversity but moderate gender fairness; SDXL 1.1 achieved balanced gender but weaker racial diversity; FLUX.1-dev exhibited strong gender bias; Kandinsky 3.0 showed both racial and gender imbalance; and QWEN-Image displayed the highest bias overall.

These findings underscore the importance of integrating systematic bias auditing into AI governance frameworks to ensure the responsible deployment of GenAI in domains such as media, education, recruitment, and healthcare. Future research should address additional demographic attributes (e.g., age, disability, non-binary identity, and Indigenous race), richer prompts, and expand research with more models. Moreover, interdisciplinary collaboration will be vital to developing GenAI systems that respect human diversity while advancing accountability, transparency, and fairness in AI innovation.

\section*{Acknowledgment}
This research was partially supported by the project Future Artificial Intelligence Research, FAIR CUP B53C220036 30006, grant number PE0000013, financed through NextGenerationEU. \\ \\
The work of Hanan Salam is also supported in part by the NYUAD Center for Interdisciplinary Data Science \& AI, funded by Tamkeen under the NYUAD Research Institute Award CG016. \\ \\
The authors thank David Gbenga Oke (Bowen University, Nigeria), Farhana Sultana (University of Gour Banga, India), Anirudha Ghosh (Visva-Bharati, India), Rim Missaoui (University of Monastir, Tunisia), and André Araújo (Universidade Federal Fluminense, Brazil) for their valuable participation with the authors of the paper in annotating the generated images. \\ \\
The authors also thank Arturo Argentieri from CNR-ISASI, Italy, for his technical contributions to the multi-GPU computing facilities.

%
%
\bibliographystyle{splncs04}

\bibliography{Ourbib}

\end{document}